\documentclass[twoside,11pt]{article}
\usepackage{jmlr2e}
\usepackage[utf8]{inputenc}
\usepackage[T1]{fontenc}
\usepackage{lmodern}
\usepackage{hyperref}
\usepackage{url}
\usepackage{booktabs}
\usepackage{soul}
\usepackage{amsfonts}
\usepackage{nicefrac}
\usepackage{microtype}
\usepackage{bbm}
\usepackage{mathtools}
\usepackage{makecell}

\title{Predicting Generalization in Deep Learning via Metric Learning -- PGDL Shared task}

\author{\name Sebastian Me\v{z}nar \email sebastjan.meznar@ijs.si \\
       \addr Jo\v{z}ef Stefan Institute\\
       Jamova 39, 1000 Ljubljana, Slovenia
       \AND
       \name Bla\v{z} \v{S}krlj \email blaz.skrlj@ijs.si   \\
       \addr Jo\v{z}ef Stefan Institute\\
       Jamova 39, 1000 Ljubljana, Slovenia \\\
       \addr Jo\v{z}ef Stefan International Postgraduate School \\
       Jamova 39,1000 Ljubljana, Slovenia}

\editor{Sebastian Me\v{z}nar}

\begin{document}

\maketitle

\begin{abstract}
    The competition ``Predicting Generalization in Deep Learning (PGDL)'' aims to provide a platform for rigorous study of generalization of deep learning models and offer insight into the progress of understanding and explaining these models. This report presents the solution that was submitted by the user \emph{smeznar} which achieved the eight place in the competition. In the proposed approach, we create simple metrics and find their best combination with automatic testing on the provided dataset, exploring how combinations of various properties of the input neural network architectures can be used for the prediction of their generalization.
\end{abstract}

\section{Introduction}

Generalization is a difficult problem in deep learning since neural models are a complex, not directly interpretable branch of models. Due to a large amount of parameters in neural network models, training objective has commonly multiple global minima that all minimize the error on the training set well, but can fail to generalize well on new data. Because of this property, choosing the right minima is crucial to achieve great performance across all data~\cite{neyshabur2017exploring}.

Recently, an extensive study of complexity measures for generalization was conducted, where authors trained over 10{,}000 convolutional networks and tested more than 40 complexity measures~\cite{jiang2020fantastic}. During testing, they found that PAC-Bayesian bounds are a promising direction for solving generalization. Another relevant observation is that the evaluation of generalization complexity measures can be misleading when the number of studied models is small.
When considering how instances differ from one another, the notion of \emph{distance} between them becomes of crucial relevance. This work also builds on some of the key ideas of \emph{metric learning}~\cite{li2018survey}, i.e. the process of automatically determining the appropriate distance score for a given problem.

The goal of the competition is to create a complexity measure that scores how well a model generalizes. The competition scores complexity measures based on the correlation between the ranking obtained from the complexity measure and the ground-truth ranking obtained empirically by ranking models based on their generalization gap. The generalization gap is defined as
\begin{equation}
g = \bigg |\frac{1}{|X|}\sum_{x_i \in X}\mathbbm{1}[f(x_i)=y_i] - \frac{1}{|\overline{X}|}\sum_{\overline{x}_i\in\overline{X}}\mathbbm{1}[f(\overline{x_i})=\overline{y_i}] \bigg |,
\end{equation}
where $x_i\in X$ is the training data, $\overline{x}_i\in\overline{X}$ is the validation data, $f$ is the classification model and $y$ and $\overline{y}$ are the ground-truth values of training and validation data~\cite{pgdl2020doc}. The metric used for calculating the correlation between two rankings is the Kendall’s Rank-Correlation Coefficient~\cite{kendall}. This coefficient is calculated on pairs representing the same rank using formula:
\begin{equation}
    \tau = \frac{\text{concordant pairs}-\text{discordant pairs}}{{n \choose 2}}.
    \label{eq:krcc}
\end{equation} 
Two pairs of points ($x_i, y_i$) and ($x_j, y_j$), where $i<j$ are said to be concordant if $x_i < x_j$ and $y_i < y_j$ holds or $x_i > x_j$ and $y_i > y_j$ holds, otherwise two pairs are discordant.

In our work we investigated whether, because of the complexity of deep learning models, there does not exist a complexity measure that exploits only some part of a model structure and gives a very accurate generalization score. Instead we focus on finding a collection of different metrics, that, when combined, give an accurate approximation of the generalization. For this we create simple metrics that represent different aspects of the model (and the data set) structure and combine them into the final approximation. Since this combination might not be trivial, we explore the combination space \emph{automatically}.

\section{Methodology}
We created seven simple metrics that give a score between 0 and 1. These metrics are the following:
\begin{itemize}
    \item \textbf{Non-uniform weights}: The percentage of weights that changed for more than the standard deviation of the initial weights. This can be defined as $$c = \frac{1}{|W|}\sum_{w\in W} \mathbbm{1}[|w_i^t - w_i| > \beta]; \beta = \sigma(\{|w_i^t - w_i|; w_i\in W\})\textbf,$$ where $W$ is a list of weights, $w_i^t$ is the trained weight $w_i$ and $\sigma(A)$ is the standard deviation of $A$.
    \item \textbf{Normalized difference}: We take the absolute difference between the end weights and the initial weights. We then divide these values by the maximum and take the mean value. This can be defined as $$c=\frac{1}{|W| \cdot \tau}\cdot\sum_{w_i\in W}|w_i-w_i^t|; \tau = \max\limits_{w_i\in W}(|w_i-w_i^t|),$$
    where $W$ is a list of weights and $w_i^t$ is the trained weight $w_i$.
    \item \textbf{Distance to a simple classifier}: The difference between the accuracy of the deep learning model and the logistic regression model on the training set. This can be defined as
    $$c=\frac{1}{|X|}\bigg|\sum_{x_i \in X}\mathbbm{1}[\textrm{DNN}(x_i)=y_i] - \sum_{x_i \in X}\mathbbm{1}[\textrm{LR}(x_i)=y_i]\bigg|$$ where $X$ is the training set, $y_i$ is the ground truth of $x_i$, $\textrm{DNN}(x_i)$ is the prediction of our deep learning model and $\textrm{LR}(x_i)$ is the prediction of our logistic regression model.
    \item \textbf{Margin}: Percentage of predictions where the difference between the two classes with the highest score is greater than $0.2$. This can be defined as $$c=\frac{1}{|X|}\cdot\sum_{x_i\in X}\mathbbm{1}[(\textrm{DNN}_1(x)-\textrm{DNN}_2(x))\geq 0.2],$$ where $X$ is the training data and $\textrm{DNN}_1(x_i), \textrm{DNN}_2(x_i)$ the prediction scores for the best and second best class for instance $x_i$.
    \item \textbf{Learning rate mean}: We train the model with different learning rates (0.0001, 0.001, 0.01) three times for five epochs and store accuracies we get after each epoch. We take the variance of accuracies for each different value of the learning rate and divide it by 0.002 (we assume variance is less than this). The final score is the mean value of these variances. This is defined as $$c=\frac{1}{3\cdot 2 \cdot 10^{-3}} \cdot \sum_{i=1}^3 \sigma^2 (\textrm{lr}_i),$$ where $\textrm{lr}_i$ are the accuracy scores for the $i$-th learning rate tested and $\sigma^2 (A)$ is the variance of $A$.
    \item \textbf{Learning rate distance}: We train the model with different learning rates (0.0001, 0.001, 0.01) three times for five epochs and store accuracies we get after each epoch. We take the variance of accuracies for each different value of the learning rate and divide it by 0.002 (we assume variance is less than this). The final score is the difference between the variances and the mean value of variances normalized. This is defined as $$c=\frac{2}{3\cdot 2 \cdot 10^{-3} } \cdot \sum_{i=1}^3 \big |\sigma^2 (\textrm{lr}_i) - \gamma \big |; \gamma =  \frac{1}{3}\sum_{j=1}^3(\sigma^2(\textrm{lr}_j)),$$ where $\textrm{lr}_i$ are the accuracy scores for the $i$-th learning rate tested and $\sigma^2 (A)$ is the variance of $A$.
    \item \textbf{Categorical cross entropy}: The categorical cross entropy of predictions on the training set. This is defined as $$c=-\frac{1}{|X|}\cdot\sum_{i=1}^{|X|}\sum_{j=1}^C y_{i, j} \cdot\log~\hat y_{i,j},$$ where $X$ is the training set, $C$ is the number of classes, $y_{i,j}$ is the ground-truth of the prediction $\hat y_{i,j}$.
\end{itemize}

We created the the prediction by summing the transformed scores of this metrics; $h(x, f)=\sum_{i=i}^7 t(x_i, f_i)$, where $f \in \{-1,0,1\}^7$ is a vector of metric inclusion and $t$ is the transformation function shown in the Equation~\ref{eq:t}.

\begin{equation}
    t(x_i, f_i) =
    \begin{cases*}
      x_i     & if $f_i = 1$  \\
      1-x_i   & if $f_i = -1$ \\
      0     & if $f_i = 0$
    \end{cases*}
    \label{eq:t}
\end{equation}

This gives us $3^7= 2187$ different combinations of simple metrics. We chose the best combination by automatically testing them on the provided dataset and choosing the one with the highest score. In our tests, the highest score was achieved by the combination where metrics "Learning rate distance" and "Non-uniform weights" were given the metric inclusion value $1$, "Learning rate mean" feature inclusion value $-1$ and all others feature inclusion value $0$.

By observing the scores of the combinations, we saw that those achieving the best results were very similar. This gives us further confidence that the chosen combination of metrics gives some insight into the model's structure and how it generalizes.

\section{Conclusion}
We introduced our methodology where simple metrics are combined to generate a complexity score. Because this combination might not be trivial, different combinations are tested automatically and the best one is chosen. Because of this, our approach might be a good starting point for creating a good method for predicting generalization.

During testing we observed that similar combination usually score similarly. For each metric we can average scores where metric inclusion value is the same. By comparing these values we can observe how metrics rank and which inclusion value is the best for a given metric.

The presented methodology has a few advantages. One of these is the modality. New metrics can easily be added or removed to improve the performance and better predict generalization. The automatic approach makes it possible to explore non-trivial combinations of simple metrics and rank them based on the overall contribution.

The main disadvantage of this approach is that we need to incorporate already existing metrics into it. The way we implement how scores from different metrics are merged also limits the possible metrics that can be added, since they need to give a score between 0 and 1, otherwise one metric would give more to the final value than the others. This can probably be avoided by using hyperparameter tuning approaches such Bayesian Optimization. Another disadvantage of our approach is that adding new metrics without discarding others might greatly impact time needed to calculate the best combination.

In further work, we want to add new metrics and discard some that did not perform well. We also want to improve automatic search for the combination of metrics using more advanced methods for parameter tuning such as Bayesian Optimization or genetic algorithms.

\bibliography{ms.bib}{}
\bibliographystyle{plain}

\end{document}